\documentclass{article}
\usepackage{ijcai20}

\usepackage{times}
\usepackage{soul}
\usepackage{url}
\usepackage[hidelinks]{hyperref}
\usepackage[utf8]{inputenc}
\usepackage[small]{caption}
\usepackage{graphicx}
\usepackage{pdfpages}
\usepackage{amsmath}
\usepackage{amsthm}
\usepackage{booktabs}
\usepackage{algorithm}
\usepackage{algorithmic}
\usepackage{multirow}
\usepackage{subfigure}

\title{End-to-end Emotion-Cause Pair Extraction via Learning to Link}

\author{
Haolin Song$^{1}$
\and
Chen Zhang$^{1}$\and
Qiuchi Li$^{2}$\And
Dawei Song$^{1,*}$
\affiliations
$^1$Beijing Institute of Technology, Beijing, China\\
$^2$University of Padua, Padua, Italy\\
$^* $ Corresponding author
\emails
\{hlsong,czhang,dwsong\}@bit.edu.cn,
qiuchili@dei.unipd.it
}

\begin{document}

\maketitle

\begin{abstract}
Emotion-cause pair extraction (ECPE), as an emergent natural language processing task, aims at jointly investigating emotions and their underlying causes in documents. It extends the previous emotion cause extraction (ECE) task, yet without requiring a set of pre-given emotion clauses as in ECE. Existing approaches to ECPE generally adopt a two-stage method, i.e., (1) emotion and cause detection, and then (2) pairing the detected emotions and causes.  Such pipeline method, while intuitive, suffers from two critical issues, including error propagation across stages that may hinder the effectiveness, and high computational cost that would limit the practical application of the method. To tackle these issues, we propose a multi-task learning model that can extract emotions, causes and emotion-cause pairs simultaneously in an end-to-end manner. Specifically, our model regards pair extraction as a link prediction task, and learns to link from emotion clauses to cause clauses, i.e., the links are directional. Emotion extraction and cause extraction are incorporated into the model as auxiliary tasks, which further boost the pair extraction. Experiments are conducted on an ECPE benchmarking dataset. The results show that our proposed model outperforms a range of state-of-the-art approaches in terms of both effectiveness and efficiency.\footnote{Code is available at \url{https://github.com/shl5133/E2EECPE}.}
\end{abstract}

\section{Introduction}

\begin{figure}
    \centering
    \subfigure[An example document and extracted emotion-cause pairs]{
    \label{fig1a}
    \includegraphics[width=0.4\textwidth]{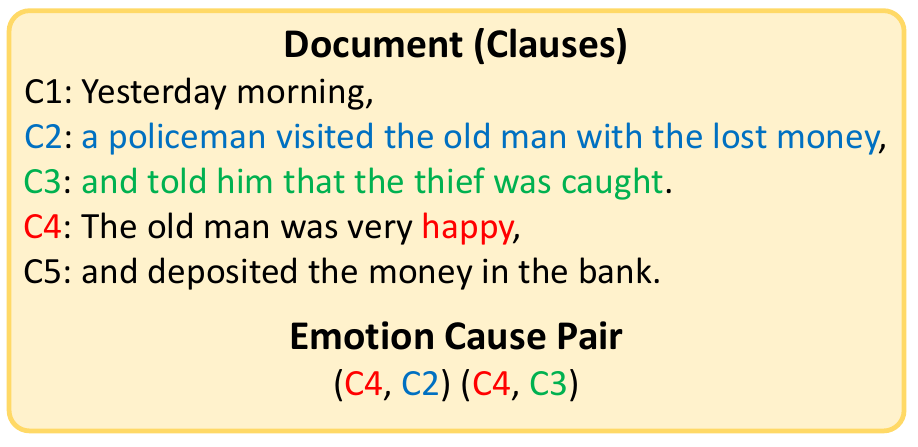}
    }
    \subfigure[Directional graph of the document]{ 
    \label{fig1b}
    \includegraphics[width=0.4\textwidth]{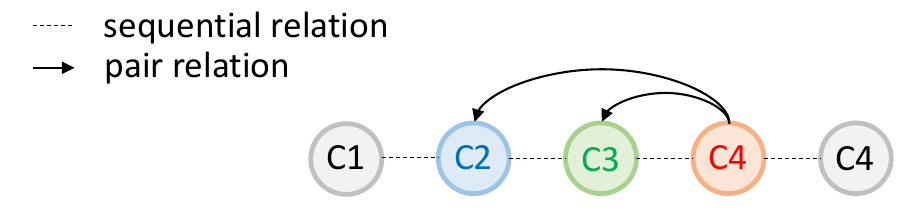}
    }
    \caption{Extracting emotion-cause pairs of the given document via learning to link.}
    \label{fig1}
\end{figure}

Emotion cause extraction (ECE)~\cite{lee2010text} aims to extract possible causes for a given emotion clause. While ECE has attracted an increasing attention due to its theoretical and practical significance, it requires that the emotion signals should be given. In practice emotion annotation is rather labor intensive, limiting the applicability of ECE in practical settings. To address the limitation of ECE, the emotion-cause pair extraction (ECPE) task was recently proposed in~\cite{xia-ding-2019-emotion}. Unlike ECE~\cite{lee2010text}, ECPE aims to extract emotions and causes without any given emotion signals, and thus better aligns with real-world applications. An illustrative example is given in Figure~\ref{fig1a}, showing that clause 4 serves as the emotion and clauses 2 and 3 are the corresponding causes. Typically, ECPE is formulated as extracting emotion-cause pairs, e.g., (clause 4, clause 2) and (clause 4, clause 3), directly from provided documents.

ECPE is a challenging task as it requires the extraction of emotions, causes and emotion-cause pairs. Existing work in ECPE mainly focuses on how to collaboratively extract emotions and causes and combine them in an appropriate way. Thus, a two-stage method~\cite{xia-ding-2019-emotion} is typically adopted, which divides pair extraction into two steps: firstly detecting emotions and causes, and then pairing them based on the likelihood of cartesian products between them. Such pipeline method approach is intuitive and straightforward. However, two critical issues arise. One is the error propagation from the first step to the second. The other issue is that, owing to the step-by-step structure, the two-stage models are often computationally expensive.

To tackle the afore-mentioned issues, we propose an end-to-end multi-task learning model for predicting emotion-cause pairs, namely \textbf{E2EECPE}, which connects emotions and causes within one single stage. More specifically, benefitting from end-to-end architectures, E2EECPE resolves the issue of error propagation while substantially reduces the computation time compared with the state-of-the-art models. Particularly, our model views emotion-cause pair extraction as a link prediction problem, as shown in Figure~\ref{fig1b}, and predicts whether there exists a directional link from the emotion to cause. Furthermore, we incorporate into the model two auxiliary tasks, namely emotion extraction and cause extraction, which are oriented to further enhance the expressiveness of the intermediate emotion representation and cause representation.

Extensive experiments are carried out on a benchmarking ECPE dataset. The experimental results demonstrate the effectiveness and efficiency of the proposed \textbf{E2EECPE} model, in comparison with a variety of state-of-the-art baselines. Moreover, further ablation study indicates that the auxiliary tasks are beneficial.
\section{Related Work}

First of all, our work is related to extracting causes based on emotions explicitly presented in documents, i.e., ECE~\cite{lee2010text}. Earlier work views ECE as a word-level sequence tagging problem and tries to solve it with corresponding tagging techniques. Therefore, primary efforts have been made on discovering refined linguistic features~\cite{chen2010emotion,lee2013detecting}, yielding improved performance. In line with other tagging related tasks such as named entity recognition (NER), support vector machines (SVMs)~\cite{gui2014emotion} and conditional random fields (CRFs)~\cite{lafferty2001conditional} have been used for ECE. More recently, instead of concentrating on word-level cause detection, clause-level extraction~\cite{gui2016event} is put forward in that the impact of individual words in a cause can span over the whole sequence in the clause.

With the emergence and development of deep representation learning, neural models has also been utilized in ECE. ~\cite{cheng2017emotion} leverages long short-term memory networks (LSTMs)~\cite{hochreiter1997long} to promote the context awareness of clause modelling. \cite{gui2017question} views the information extraction problem as the retrieval task in question answering (QA) and examines the effect of memory networks~\cite{sukhbaatar2015end} for extraction. Likewise, taking advantage of attention mechanism~\cite{bahdanau2014neural}, ~\cite{li2018co} employs a co-attention based model and achieves the state-of-the-art performance. 

In light of recent advances in multi-task learning, joint extraction of emotions and causes is investigated ~\cite{chen2018joint} to exploit the mutual information between two correlated tasks. However, these works do not explicitly combine two tasks into one. Thereafter, ~\cite{xia-ding-2019-emotion} argues that, while co-extraction of emotions and causes is important, emotion-cause pair extraction (ECPE) is a more challenging problem that is worth putting more emphasis on. Nevertheless, ~\cite{xia-ding-2019-emotion} adopts a two-stage approach, which performs emotion and cause extraction first and then pairs the extracted emotions and causes. As discussed in the previous section, such two stage approach suffers from error propagation and high computation cost.


Our work aims to tackle these challenges in ECPE. Rather than processing emotion-cause pair extraction as a two stage task (as used in the existing work), we consolidate two stages into a unified multi-task learning framework, and further consider it as a link prediction task which could be solved in an end-to-end manner.
\section{Methodology}

To solve ECPE in end-to-end fashion, we take inspiration from the link prediction problem in graph learning, which aims at predicting potential edges between unconnected vertices in a graph. Essentially, if we consider emotion-cause pairs in a document as triplets in a graph, then the extraction of such pairs is a sort of link prediction from a graph that is at first armed with no edges but only vertices. In order to achieve above procedure, we borrow the idea of learning a graph-based dependency parser~\cite{dozat2016deep} and adapt it to our target task. Coupling link prediction with auxiliary emotion extraction and cause extraction tasks, our model is capable of jointly, and more effectively, extracting emotions, causes, and emotion-cause pairs. 

\begin{figure}
    \centering
    \includegraphics[width=0.47\textwidth]{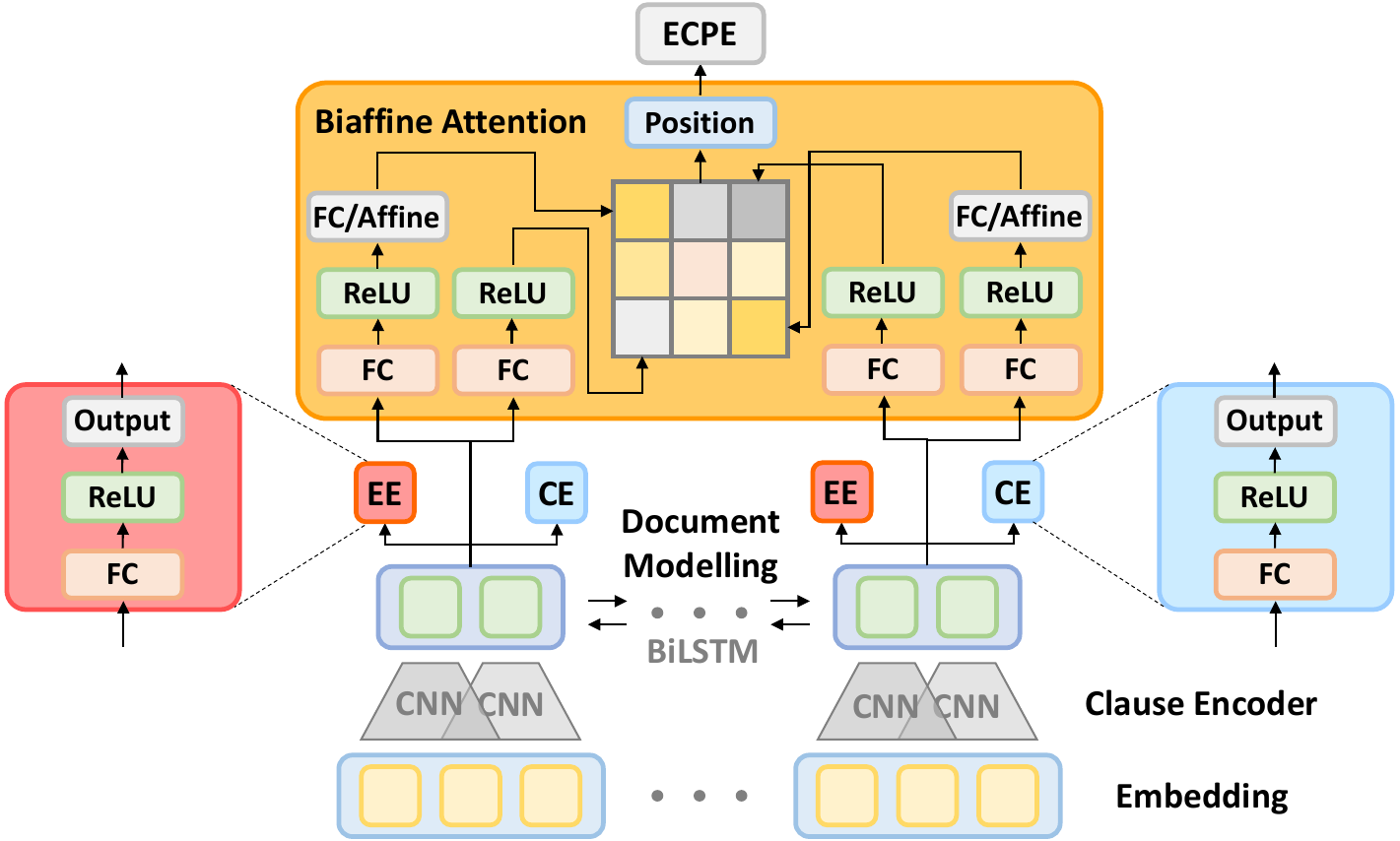}
    \caption{An overview of \textbf{E2EECPE}. EE, CE are short for emotion extraction, cause extraction respectively.}
    \label{fig2}
\end{figure}

\subsection{Problem Formulation}

Generally, for any provided document $D$, we could split it into a sequence of clauses based on punctuation, i.e., $D=\{c_i\}_{i=1}^{|D|}$, where $c_i$ could further be decomposed into words, i.e., $c_i=\{w_j\}_{j=1}^{|c_i|}$. Here, $|D|$ is the number of clauses in the document and $|c_i|$ is the number of words in the $i$-th clause. ECPE aims to extract a set of $|P|$ emotion-cause pairs $P=\{(c_k^e,c_k^c)\}_{k=1}^{|P|}$ from the document $D$, where  $c_k^e$, $c_k^c$ represents the emotion clause and the cause clause in the $k$-th pair, respectively.

\subsection{Overall Architecture}

An overview of the proposed \textbf{E2EECPE} approach is shown in Figure~\ref{fig2}. The bottom layer is a clause encoder (Section~\ref{sec3.3}) and a document modelling layer (Section~\ref{sec3.4}) which transform the word embeddings into the contextualized clause representations. The middle part consists of auxiliary tasks (Section~\ref{sec3.7}), i.e., emotion extraction and cause extraction. The top most part is a biaffine attention layer (Section~\ref{sec3.5}) which first encodes interaction between the emotion representation and cause representation, and then outputs a postion-weighted pair matrix for pair extraction.

\subsection{Embedding \& Clause Encoder}

\label{sec3.3}

With the purpose of integrating words into clause-level neural models, we embed each word in a clause into low-dimensional vectors~\cite{bengio2003neural}, by which we could represent each word in the clause with its vector representation\footnote{If not specified, we use notations in bold as the vector representations of their original concepts.} $c_i=\{\mathbf{w}_j\}_{j=1}^{|c_i|}$, where $\mathbf{w}_j\in\mathbf{R}^{d_e}$ and $d_e$ is the dimensionality of the embedding. 

After that, we need to attain contextualized representations of clauses. Owing to the recognized performance and local context awareness of the convolutional neural networks (CNNs) on text classification benchmarks~\cite{kim2014convolutional}, we adopt CNNs as the backbone of our clause encoder. 

For an embedded clause $c=\{\mathbf{w}_j\}_{j=1}^{|c|}$, we apply one-dimensional convolution operations with kernels
of different sizes over the word sequence:
\begin{equation}
    \mathbf{C}_t=\sigma(\mathrm{conv}_t(\mathbf{w}_1,\cdots,\mathbf{w}_{|c|}))
\end{equation}

\noindent where $\mathrm{conv}_t$ denotes the $t$-th convolution operation and $\mathbf{C}_t\in\mathbf{R}^{|c|\times d_c}$ is the output of the operation. $d_c$ is the number of filters employed in one convolution operation and $\sigma(\cdot)$ used here is actually $\mathrm{max}(0, \cdot)$. 

Then max-pooling is used to distill the features for concatenation. Hence, we finally get context-aware features for the clause:
\begin{gather}
    \tilde{\mathbf{c}}_t=\mathrm{maxpool}(\mathbf{C}_t) \\
    \mathbf{c}=[\oplus_{t}\tilde{\mathbf{c}}_t]
\end{gather}

\noindent where $\mathbf{c}\in\mathbf{R}^{|t|\cdot d_c}$ is the convoluted feature and $\oplus$ means vector concatenation. $|t|$ is the total number of convolution operations.

\subsection{Document Modelling}

\label{sec3.4}

Since we have sequential clauses in a document, the influences brought by document-level structures become a crucial part that we should fit into our model. A straightforward idea is to leverage temporal relations among clauses with LSTMs. 

Specifically, provided with the encoded clause representations $\{\mathbf{c}_i\}_{i=1}^{|c|}$, we employ a bidirectional LSTM to update clause-level features and get $\mathbf{h}_i \in\mathbf{R}^{2d_h}$:
\begin{equation}
    \mathbf{h}_i=[\mathrm{LSTM}_f(\mathbf{c}_i)\oplus\mathrm{LSTM}_b(\mathbf{c}_i)]
\end{equation}

\noindent where $\mathrm{LSTM}_f(\cdot)$ and $\mathrm{LSTM}_b(\cdot)$ denote the forward and backward unidirectional LSTMs, respectively. $d_h$ is the dimensionality of hidden states for a unidirectional LSTM.

\subsection{Biaffine Attention}

\label{sec3.5}

Motivated by advances in link prediction~\cite{schlichtkrull2018modeling}, we can directly compute the similarity scores among vertex representations (clause representations in our task), e.g., $\sigma(\mathbf{z}_p^{\top} \mathbf{z}_q)$ for any representations of vertex $p$ and $q$, to make predictions. However, the above predictions are only concerned with undirectional circumstances since $\mathbf{z}_p^{\top} \mathbf{z}_q=\mathbf{z}_q^{\top} \mathbf{z}_p$, which is not adequate for emotion-cause pair extraction. To solve the problem, we utilize biaffine transform to complete the filling of adjacent matrices, which are called \emph{pair matrices} in our work. This idea is similar to dependency parsing~\cite{dozat2016deep} that is also directional.

\subsubsection{Emotion \& Cause Representation}

According to biaffine attention mechanism, each vertex in the graph should have two independent representations, i.e., one is for pointing out and the other for pointed in. In doing so, the pair matrix output by the transformation is asymmetric and direction-aware.

The emotion representation and cause representation are separately offered as below:
\begin{gather}
    \mathbf{z}_i^e=\sigma(\mathbf{W}^e\mathbf{h}_i+\mathbf{b}^e) \\
    \mathbf{z}_i^c=\sigma(\mathbf{W}^c\mathbf{h}_i+\mathbf{b}^c)
\end{gather}

\noindent where $\mathbf{W}^e\in\mathbf{R}^{d_z\times 2d_h}$, $\mathbf{b}^e\in\mathbf{R}^{d_z}$ and $\mathbf{W}^c\in\mathbf{R}^{d_z\times 2d_h}$, $\mathbf{b}^c\in\mathbf{R}^{d_z}$ are two sets of trainable weights and biases, respectively for the emotion and cause representations.

\subsubsection{Biaffine Transform}

Then, we implement biaffine transform on the collected emotion and cause representations. In other words, with the purpose of merging these two kinds of representations into our aimed pair matrices, we fold emotion-cause dynamics into two components. On the one hand, we need to perform a bilinear like operation on each possible pair of emotion and cause. On the other hand, we believe bilinear transform is not enough to deal with such complicated interactions, and thus we facilitate it by injecting bias.

More specifically, we calculate each entry in the expected pair matrix as follows:
\begin{equation}
    \mathbf{M}_{p,q}=(\mathbf{W}^{m}\mathbf{z}_p^e+\mathbf{b}^m)^{\top}\mathbf{z}_q^c
\end{equation}

\noindent where $\mathbf{W}^m \in\mathbf{R}^{d_z\times d_z}$ and $\mathbf{b}^m \in\mathbf{R}^{d_z}$ are learnable parameters of affine transform, while $\mathbf{M}_{p,q}$ indicates an entry of the pair matrix in the $p$-th row, $q$-th column.

Constrained by the inherent property of an adjacent matrix, we further activate the pair matrix with the sigmoid function $g(\cdot)$:
\begin{equation}
    \tilde{\mathbf{M}}_{p,q}=g(\mathbf{M}_{i,j})
\end{equation}

\subsection{Asymmetric Position Weight Matrix}

A trivial observation on the co-occurrence patterns of emotions and causes is that the emotion and the cause in a unique pair appear near each other in term of their absolute positions in the document. Thus, position embeddings are introduced to directly encode positions into vectors~\cite{xia-ding-2019-emotion}. Different from the existing approaches, our work is based on graph learning, and thereby can not be aided by manipulation of embeddings. Instead, we apply proximity weights on features as in~\cite{zhang2019syntax}, but extent it to matrices.

Moreover, we notice that in reality people are more likely to inform the causes before expressing emotions, which will be verified with statistics of dataset, implying we should assign asymmetric position weight matrices, instead of symmetric ones, to the original matrix representations.


Specifically, the entries in the lower triangular of position weight matrix for the given document is of more significance:
\begin{equation}
    \mathbf{A}_{p,q}=\frac{|D|-|p-q-1|+\epsilon}{|D|+\epsilon}
\end{equation}

\noindent where $\epsilon$ is a small number for smoothing.

Finally, we obtain the features for indicating pair links as follows:
\begin{equation}
    \hat{\mathbf{M}}_{p,q}=\tilde{\mathbf{M}}_{p,q}\odot\mathbf{A}_{p,q}
\end{equation}

\noindent where $\odot$ denotes element-wise multiplication.

\subsection{Multi-task Setting}

\label{sec3.7}

As aforementioned, emotion extraction and cause extraction can be viewed as auxiliary tasks to augment the emotion and cause representations for constructing more expressive pair matrix. Hence we develop a multi-task paradigm, which shares the fundamental part of network structure for the main task with auxiliary tasks. To achieve this goal, we first acquire features dedicated for classification with following procedure: 
\begin{gather}
    \tilde{\mathbf{z}}_i^e=\sigma(\tilde{\mathbf{W}}^e\mathbf{h}_i+\tilde{\mathbf{b}}^e) \\
    \tilde{\mathbf{z}}_i^c=\sigma(\tilde{\mathbf{W}}^c\mathbf{h}_i+\tilde{\mathbf{b}}^c)
\end{gather}

\noindent where $\tilde{\mathbf{W}}^e\in\mathbf{R}^{d_z\times 2d_h}$, $\tilde{\mathbf{b}}^e\in\mathbf{R}^{d_z}$ and $\tilde{\mathbf{W}}^c\in\mathbf{R}^{d_z\times 2d_h}$, $\tilde{\mathbf{b}}^c\in\mathbf{R}^{d_z}$ are again two sets of trainable weights and biases, respectively. 

Subsequently, predictions are produced by two fully connected layers followed by softmax normalization layers:
\begin{gather}
    \hat{\mathbf{y}}^e=\mathrm{softmax}(\hat{\mathbf{W}}^e\tilde{\mathbf{z}}_i^e+\hat{\mathbf{b}}^e) \\
    \hat{\mathbf{y}}^c=\mathrm{softmax}(\hat{\mathbf{W}}^c\tilde{\mathbf{z}}_i^c+\hat{\mathbf{b}}^c)
\end{gather}

\noindent where $\hat{\mathbf{W}}^e\in\mathbf{R}^{2\times d_z}$, $\hat{\mathbf{b}}^e\in\mathbf{R}^{2}$ and $\hat{\mathbf{W}}^c\in\mathbf{R}^{2\times d_z}$, $\hat{\mathbf{b}}^c\in\mathbf{R}^{2}$ are weights and biases for learning. 

\subsection{Training Objective}

Eventually, the whole structure can be trained by standard gradient descent. Accordingly, the objective function is a combination of cross entropy with $L_2$-norm regularization, formulated as below:
\begin{gather}
    \begin{split}
    \mathcal{L}_{pair}=-\sum_{p,q}\mathbf{Y}_{p,q}\mathrm{log}(\hat{\mathbf{M}}_{p,q}) \\
    -\sum_{p,q}(1-\mathbf{Y}_{p,q})\mathrm{log}(1-\hat{\mathbf{M}}_{p,q})
    \end{split} \\
    \mathcal{L}_{aux}=-\sum_{i}[\sum_k\mathbf{y}^e_{k}\mathrm{log}(\hat{\mathbf{y}}^e_{k})+\sum_k\mathbf{y}^c_{k}\mathrm{log}(\hat{\mathbf{y}}^c_{k})]
\end{gather}

\noindent where $(p, q)$ and $i$, $k$ serve as enumerators over all elements. $\mathbf{y}^e$, $\mathbf{y}^c$, and $\mathbf{Y}$ are correspondingly the ground truth.

Furthermore, we add two coefficients to balance the influences of above two objective functions. The ultimate training objective then becomes:
\begin{equation}
    \mathcal{L}=\mathcal{L}_{pair}+\beta\mathcal{L}_{aux}+\lambda||\theta||_2
\end{equation}

\noindent where the term $\beta$ is used to adjust the potential influences brought by multi-task learning, which is refined according to a pilot study. $\theta$ stands for all parameters that need to be optimized, while $\lambda$ is a coefficient for $L_2$-norm regularization.

\subsection{Inference}

With well trained model, we can infer emotion-cause pairs by comparing each entry in $\hat{\mathbf{M}}$ with a predefined threshold $\eta$
\begin{equation}
    \hat{\mathbf{Y}}_{p,q}=
    \begin{cases}
    1, &\hat{\mathbf{M}}_{p,q}>\eta \\
    0, &\hat{\mathbf{M}}_{p,q}\leq\eta
    \end{cases}
\end{equation}

\noindent where $\hat{\mathbf{Y}}$ is the inference result matrix with binary (1-0) indicators.
\section{Experimental Setting}

\subsection{Dataset}

We carry out experiments on a publicly available dataset for emotion-cause pair extraction, released by~\cite{xia-ding-2019-emotion}. Consisting of news crawled from web, the dataset is referred to as \textsc{News} in the rest of the paper. It is randomly split into ten folds. In our experiments, to evaluate our trained model, each fold is further divided into two parts, namely train set and test set which respectively take 90\% and 10\% of the data. Table~\ref{tab1} shows some basic statistics of the dataset. A key observation is that most documents only contain one emotion-cause pair therein, implying the sparsity of the pair matrix. Therefore the issue of label imbalance will be elaborated in following discussions. Moreover, a large amount of emotion-cause pairs have the emotion and the cause within 1 relative offset, suggesting the necessity of using proximity constraints (exactly what position weight matrix does) in the predicted pair matrix.  

\begin{table}
\centering
\begin{tabular}{lr}
\toprule
  & \textsc{News}  \\
\midrule
\# of documents                                       & 2105         \\
avg. \# of clauses per document                & 14           \\
\# of EC pairs                             & 2167         \\
\# of documents with 1 EC pair          & 1746         \\
\# of documents with over 1 EC pairs  & 421          \\
\midrule
\# of EC pairs with 0 relative offset                      & 511           \\
\# of EC pairs with 1 relative offset                      & 1342          \\
\# of EC pairs with 2 relative offset                      & 224           \\
\# of EC pairs with over 2 relative offset                 & 90            \\
maximum EC pair offset                & 12            \\
avg. offset of EC pairs    & 0.9977        \\
\bottomrule
\end{tabular}
\caption{Statistics of the dataset. EC stands for emotion-cause, and relative offset indicates the absolute distance between the emotion and the cause of a pair in the document.}
\label{tab1}
\end{table}

\subsection{Parameter Settings}

For all our experiments, pre-trained word vectors on Weibo (a Chinese micro-bloging website) using Word2Vec~\cite{mikolov2013efficient} are leveraged to initialize the word embeddings. The dimensionality
of embeddings (i.e., $d_e$) is set to 200. We use 4 convolutional layers (i.e., $|t|$) whose kernel sizes are \{2,3,4,5\} for the clause encoder and the number of filters for all the convolutional layers (i.e., $d_c$) is 50, for capturing
gram-level features. In order to avoid overfitting, we apply dropout to embeddings and outputs of the clause encoder, yielding 0.5 probability of randomized zeroes on features. The dimensionality for hidden states of a unidirectional LSTM (i.e., $d_h$) is 300. The dimensionalities for all fully connected layers in the main task and auxiliary tasks (i.e., $d_z$) are 100. Moreover, the batch size and learning rate are determined through grid parameter search, which are 32 and 10\textsuperscript{-3}, respectively. The coefficient for $L_2$-norm regularization (i.e., $\lambda$) is 10\textsuperscript{-5} . Based on a pilot study, we find the best value for the threshold (i.e., $\eta$) in the inference stage is 0.3, which will be detailed in next section. The coefficient for the trade-off in objective function (i.e., $\beta$) is 1. In addition, the smoothing term in the calculation of position weight matrix (i.e., $\epsilon$) is 1. Furthermore, Adam is used as the optimizer and all trainable parameters are randomly initialized with uniform distribution~\cite{he2015delving}.

\subsection{Baselines \& Evaluation Metrics}

Our approach\footnote{Code is available in \url{https://github.com/shl5133/E2EECPE}.} is compared with a range of strong baselines, which are the state-of-the-art methods proposed by~\cite{xia-ding-2019-emotion} for emotion-cause pair extraction. These baselines are either one-stage or two-stage models.

The two-stage models are listed below. They first extract emotions and causes with multi-task architectures independently or interactively, then classify the cartesian products of emotions and causes extracted in the first stage into pairs or non-pairs. 

\begin{itemize}
    \item \textbf{Indep} firstly considers emotion extraction and cause extraction as independent tasks and extract emotions and causes with multi-task learning, then pairs the extracted emotions and causes with a classifier.
    \item \textbf{Inter-CE} and \textbf{Inter-EC} typically follow the procedure of \textbf{Indep}, however, assist emotion extraction and cause extraction with directed interaction modelling.
\end{itemize}

The one-stage models neglect the second stage in the two-stage models, and consider the cartesian products as predictions. They are listed as follows. 

\begin{itemize}
    \item \textbf{Indep} w/o filter removes the classifier of \textbf{Indep}.
    \item \textbf{Inter-CE} w/o filter removes the classifier of \textbf{Inter-CE}.
    \item \textbf{Inter-EC} w/o filter removes the classifier of \textbf{Inter-EC}.
\end{itemize}

Precision, recall, and macro F1 measures are adopted as effectiveness metrics in our experiments. Meanwhile, runtime is used as a measure of efficiency. The final results are obtained by averaging the ten folds results.

\section{Result \& Analysis}

\subsection{Results in Effectiveness}

We perform a comparison of \textbf{E2EECPE} with one-stage and two-stage baseline models to quantitatively understand in what ways \textbf{E2EECPE} is more effective than the baselines.

\begin{table*}
\centering
\begin{tabular}{lrrrrrrrrr}
\toprule
\multirow{2}{*}{Models} & \multicolumn{3}{c}{emotion extraction} & \multicolumn{3}{c}{cause extraction} & \multicolumn{3}{c}{emotion-cause pair extraction}   \\
\cmidrule(lr){2-10}
~  & P & R & F1 & P & R & F1 & P & R & F1   \\
\midrule
\textbf{Indep} w/o filter        & 0.8483 & 0.7961 & 0.8208 & 0.6898 & 0.5648 & 0.6198 & 0.5932  & 0.5094  & 0.5470      \\
\textbf{Inter-CE} w/o filter     & 0.8458 & 0.8035 & {\bf 0.8263} & 0.6838 & 0.5754 & 0.6231 & 0.5827  & 0.5300  & 0.5531     \\
\textbf{Inter-EC} w/o filter     & 0.8406 & {\bf 0.8097} & 0.8242 & 0.6989 & 0.5991 & 0.6426 & 0.5975  & 0.5538  & 0.5716     \\
\midrule
\textbf{Indep}          & 0.8483 & 0.7961 & 0.8208 & 0.6898 & 0.5648 & 0.6198 & {\bf 0.6943}  & 0.5047  & 0.5833      \\
\textbf{Inter-CE}        & 0.8458 & 0.8035 & {\bf 0.8263} & 0.6838 & 0.5754 & 0.6231 & 0.6780  & 0.5254  & 0.5896     \\
\textbf{Inter-EC}        & 0.8406 & {\bf 0.8097} & 0.8242 & 0.6989 & 0.5991 & 0.6426 & 0.6691  & 0.5503  & 0.6013     \\
\midrule
\textbf{E2EECPE}                   & {\bf 0.8595}\textsuperscript{\dag} & 0.7915 & 0.8238 & {\bf 0.7062} & {\bf 0.6030} & {\bf 0.6503} & 0.6478  & {\bf 0.6105}\textsuperscript{\dag}  & {\bf 0.6280}\textsuperscript{\dag}     \\
\bottomrule
\end{tabular}
\caption{Comparison results of emotion extraction, cause extraction, and emotion-cause pair extraction with precision, recall, and F1-measure as metrics. The results in bold are the best performing ones under each column. The results of emotion extraction and cause extraction for one-stage and two-stage models are exactly the same because one-stage models are ablated ones of two-stage models. \textsuperscript{\dag} indicates results that are significantly better than best performing baseline \textbf{Inter-EC} with paired t-test (\textit{p} is smaller than 0.05).}
\label{tab2}
\end{table*}

Table~\ref{tab2} gives the results in terms of precision, recall and macro-F1 measures. The comparison results demonstrate that our model \textbf{E2EECPE} consistently outperforms the baselines for the main task (emotion-cause pair extraction) with regard to recall and F1, indicating the representation power and the effectiveness of our model. Nevertheless, we also observe that our model performs less well in precision than the two-stage baseline models. With additional observation that the baseline models are performing poorly on recall, we conjecture the existing models suffer from predicting only few testing instances as pairs. 

Furthermore, \textbf{E2EECPE} is superior on the two auxiliary tasks (emotion extraction and cause extraction). We attribute the improvement to multi-task structure in our model which combines auxiliary tasks and the main task. Apart from that, the one-stage models yield lower results than \textbf{E2EECPE} on cause extraction, suggesting that error propagation is a comparably severe issue in the existing models but is alleviated in our model.

\subsection{Runtime Analysis}

To confirm that our end-to-end model is more efficient than the two-stage models, we perform runtime analysis among the two-stage baseline models and ours. Concretely, we display the average running time consumed by models on a epoch in different folds. 

\begin{figure}[h]
    \centering
    \includegraphics[width=0.45\textwidth]{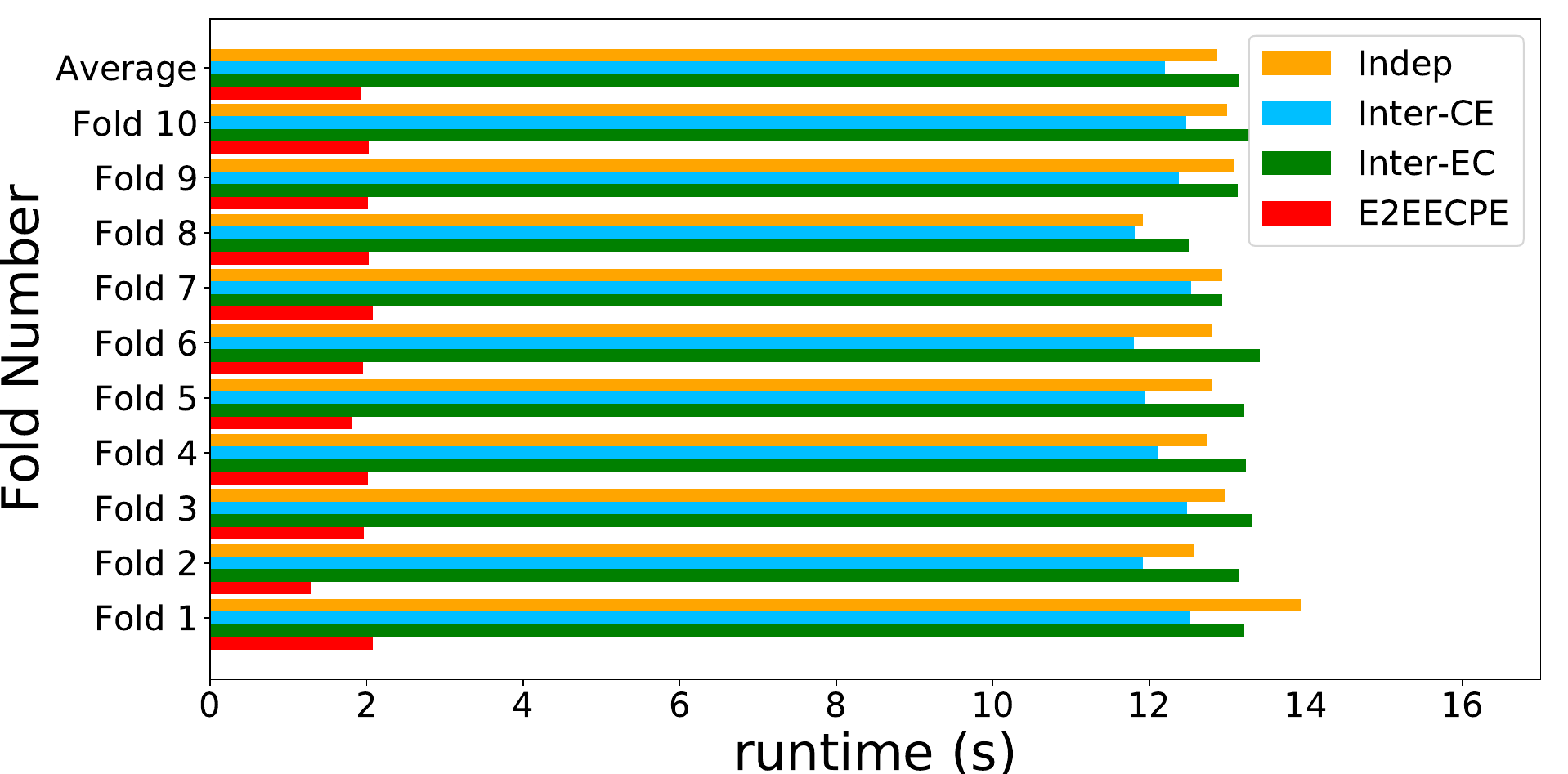}
    \caption{Runtime analysis (s).}
    \label{fig4}
\end{figure}

Results listed in Figure~\ref{fig4} suggests that our model is 6-7 times faster than two-stage models, indicating the efficiency of our end-to-end model.

\subsection{Ablation Study}

To understand the efficacy of auxiliary tasks and position weight matrix, we conduct an ablation study on \textbf{E2EECPE}. Specifically, we separately ablate auxiliary tasks and position weight matrix from \textbf{E2EECPE}, and call them \textbf{E2EECPE} \textit{w/o auxiliary} and \textbf{E2EECPE} \textit{w/o position}, respectively.

\begin{table}
\centering
\begin{tabular}{lrrr}
\toprule
Models  & P & R & F1 \\
\midrule
\textbf{E2EECPE}                              & {\bf 0.6478}  & 0.6105  & {\bf 0.6280}\\
\midrule
\textbf{E2EECPE} \textit{w/o auxiliary}                       & 0.5982  & 0.5340  & 0.5635  \\
\textbf{E2EECPE} \textit{w/o position}       & 0.6421  & {\bf 0.6158}  & 0.6275      \\
\bottomrule
\end{tabular}
\caption{Ablation study results. The results in bold are the best performing ones under each column.}
\label{tab3}
\end{table}

The results in Table~\ref{tab3} show a significant performance drop of \textbf{E2EECPE} \textit{w/o auxiliary} and a relatively minor drop of \textbf{E2EECPE} \textit{w/o position} compared with \textbf{E2EECPE},  verifying the remarkable benefit of the multi-task learning schema. Meanwhile, the results that \textbf{E2EECPE} \textit{w/o position} only differs slightly from \textbf{E2EECPE} based on all metrics, indicate the fact that imposing position information is still of importance.

\subsection{Effect of Threshold in Inference Stage}

Inference based on pair matrix is powerful, yet we do not exactly know what threshold (i.e., $\eta$) is the most suitable one for its expressiveness. It is therefore helpful to explore the effect of the threshold by altering it and examining the results.

\begin{table}
\centering
\begin{tabular}{lrrr}
\toprule
$\eta$  & P & R & F1 \\
\midrule
0.2         & 0.5743  & {\bf 0.6456}  & 0.6071      \\
0.3         & 0.6478  & 0.6105  & {\bf 0.6280}      \\
0.4         & 0.6757  & 0.5849  & 0.6265            \\
0.5         & 0.7185  & 0.5543  & 0.6255            \\
0.6         & {\bf 0.7326}  & 0.5385  & 0.6201      \\
\bottomrule
\end{tabular}
\caption{Effect of threshold. The results in bold are the best performing ones under each column.}
\label{tab4}
\end{table}

From Table~\ref{tab4}, we conclude that 0.3 is the most appropriate one for our studied task. With increases of $\eta$, drops of F1 are noted, implying potential loss of extracted pairs. In addition, we also speculate that the reason why the best value is not around 0.5 (the expectation of random variables ranging uniformly from 0 to 1) is that the element-wise multiplication of a position weight matrix with the sigmoid-activated pair matrix produces a smaller expectation (as upper bound decreases). 

\subsection{Issue of Label Imbalance}

In order to measure the impact brought by label imbalance, typically in the form of pair matrix sparsity, we remove the examples containing more than one pair for test set in each fold to make up a \texttt{Hard} dataset, then record the mean results across ten folds correspondingly. 

\begin{table}
\centering
\begin{tabular}{lrrr}
\toprule
Dataset  & P & R & F1 \\
\midrule
\texttt{Full}         & 0.6478  & 0.6105  & 0.6280      \\
\texttt{Hard}         & 0.6002  & 0.6479  & 0.6226      \\
\bottomrule
\end{tabular}
\caption{The results for verifying the issue of label imbalance. }
\label{tab5}
\end{table}

We can observe in Table~\ref{tab5} that our model encounters a failure on the \texttt{Hard} dataset with decreases on precision and F1 measure, suggesting that further investigation is needed to solve this problem.

\section{Conclusion and Future Work}

The emotion-cause pair extraction task is a new and more realistic task that seeks to identify emotion-cause pairs in documents. However, previous two-stage models are inherently limited by the idea of solving this task via two stages. To this end, we propose an end-to-end model that regards the oriented problem as predicting directional links between emotions and causes via biaffine attention. Additionally, we also aid the model with auxiliary tasks and position weight matrix. Experimental results prove the superiority of our model over other baselines.

Based on the work in this paper, we believe there are some promising directions yet to be explored. On the one hand, more fancy models such as graph neural networks are expected to be developed to incorporate with learned position information instead of refined one. On the other hand, the label imbalance issue should be addressed with task-specific tactics.


\bibliographystyle{named}
\bibliography{ijcai20}

\begin{thebibliography}{}

\bibitem[\protect\citeauthoryear{Bahdanau \bgroup \em et al.\egroup
  }{2014}]{bahdanau2014neural}
Dzmitry Bahdanau, Kyunghyun Cho, and Yoshua Bengio.
\newblock Neural machine translation by jointly learning to align and
  translate.
\newblock {\em arXiv preprint arXiv:1409.0473}, 2014.

\bibitem[\protect\citeauthoryear{Bengio \bgroup \em et al.\egroup
  }{2003}]{bengio2003neural}
Yoshua Bengio, R{\'e}jean Ducharme, Pascal Vincent, and Christian Jauvin.
\newblock A neural probabilistic language model.
\newblock {\em Journal of machine learning research}, 3(Feb):1137--1155, 2003.

\bibitem[\protect\citeauthoryear{Chen \bgroup \em et al.\egroup
  }{2010}]{chen2010emotion}
Ying Chen, Sophia Yat~Mei Lee, Shoushan Li, and Chu-Ren Huang.
\newblock Emotion cause detection with linguistic constructions.
\newblock In {\em Proceedings of the 23rd International Conference on
  Computational Linguistics}, pages 179--187. Association for Computational
  Linguistics, 2010.

\bibitem[\protect\citeauthoryear{Chen \bgroup \em et al.\egroup
  }{2018}]{chen2018joint}
Ying Chen, Wenjun Hou, Xiyao Cheng, and Shoushan Li.
\newblock Joint learning for emotion classification and emotion cause
  detection.
\newblock In {\em Proceedings of the 2018 Conference on Empirical Methods in
  Natural Language Processing}, pages 646--651, 2018.

\bibitem[\protect\citeauthoryear{Cheng \bgroup \em et al.\egroup
  }{2017}]{cheng2017emotion}
Xiyao Cheng, Ying Chen, Bixiao Cheng, Shoushan Li, and Guodong Zhou.
\newblock An emotion cause corpus for chinese microblogs with multiple-user
  structures.
\newblock {\em ACM Transactions on Asian and Low-Resource Language Information
  Processing (TALLIP)}, 17(1):6, 2017.

\bibitem[\protect\citeauthoryear{Dozat and Manning}{2016}]{dozat2016deep}
Timothy Dozat and Christopher~D Manning.
\newblock Deep biaffine attention for neural dependency parsing.
\newblock {\em arXiv preprint arXiv:1611.01734}, 2016.

\bibitem[\protect\citeauthoryear{Gui \bgroup \em et al.\egroup
  }{2014}]{gui2014emotion}
Lin Gui, Li~Yuan, Ruifeng Xu, Bin Liu, Qin Lu, and Yu~Zhou.
\newblock Emotion cause detection with linguistic construction in chinese weibo
  text.
\newblock In {\em Natural Language Processing and Chinese Computing}, pages
  457--464. Springer, 2014.

\bibitem[\protect\citeauthoryear{Gui \bgroup \em et al.\egroup
  }{2016}]{gui2016event}
Lin Gui, Dongyin Wu, Ruifeng Xu, Qin Lu, and Yu~Zhou.
\newblock Event-driven emotion cause extraction with corpus construction.
\newblock In {\em Proceedings of the 2016 Conference on Empirical Methods in
  Natural Language Processing}, pages 1639--1649, 2016.

\bibitem[\protect\citeauthoryear{Gui \bgroup \em et al.\egroup
  }{2017}]{gui2017question}
Lin Gui, Jiannan Hu, Yulan He, Ruifeng Xu, Lu~Qin, and Jiachen Du.
\newblock A question answering approach for emotion cause extraction.
\newblock In {\em Proceedings of the 2017 Conference on Empirical Methods in
  Natural Language Processing}, pages 1593--1602, 2017.

\bibitem[\protect\citeauthoryear{He \bgroup \em et al.\egroup
  }{2015}]{he2015delving}
Kaiming He, Xiangyu Zhang, Shaoqing Ren, and Jian Sun.
\newblock Delving deep into rectifiers: Surpassing human-level performance on
  imagenet classification.
\newblock In {\em Proceedings of the IEEE international conference on computer
  vision}, pages 1026--1034, 2015.

\bibitem[\protect\citeauthoryear{Hochreiter and
  Schmidhuber}{1997}]{hochreiter1997long}
Sepp Hochreiter and J{\"u}rgen Schmidhuber.
\newblock Long short-term memory.
\newblock {\em Neural computation}, 9(8):1735--1780, 1997.

\bibitem[\protect\citeauthoryear{Kim}{2014}]{kim2014convolutional}
Yoon Kim.
\newblock Convolutional neural networks for sentence classification.
\newblock In {\em Proceedings of the 2014 Conference on Empirical Methods in
  Natural Language Processing (EMNLP)}, pages 1746--1751, 2014.

\bibitem[\protect\citeauthoryear{Lafferty \bgroup \em et al.\egroup
  }{2001}]{lafferty2001conditional}
John~D Lafferty, Andrew McCallum, and Fernando~CN Pereira.
\newblock Conditional random fields: Probabilistic models for segmenting and
  labeling sequence data.
\newblock In {\em Proceedings of the Eighteenth International Conference on
  Machine Learning}, pages 282--289. Morgan Kaufmann Publishers Inc., 2001.

\bibitem[\protect\citeauthoryear{Lee \bgroup \em et al.\egroup
  }{2010}]{lee2010text}
Sophia Yat~Mei Lee, Ying Chen, and Chu-Ren Huang.
\newblock A text-driven rule-based system for emotion cause detection.
\newblock In {\em Proceedings of the NAACL HLT 2010 Workshop on Computational
  Approaches to Analysis and Generation of Emotion in Text}, pages 45--53.
  Association for Computational Linguistics, 2010.

\bibitem[\protect\citeauthoryear{Lee \bgroup \em et al.\egroup
  }{2013}]{lee2013detecting}
Sophia Yat~Mei Lee, Ying Chen, Chu-Ren Huang, and Shoushan Li.
\newblock Detecting emotion causes with a linguistic rule-based approach 1.
\newblock {\em Computational Intelligence}, 29(3):390--416, 2013.

\bibitem[\protect\citeauthoryear{Li \bgroup \em et al.\egroup
  }{2018}]{li2018co}
Xiangju Li, Kaisong Song, Shi Feng, Daling Wang, and Yifei Zhang.
\newblock A co-attention neural network model for emotion cause analysis with
  emotional context awareness.
\newblock In {\em Proceedings of the 2018 Conference on Empirical Methods in
  Natural Language Processing}, pages 4752--4757, 2018.

\bibitem[\protect\citeauthoryear{Mikolov \bgroup \em et al.\egroup
  }{2013}]{mikolov2013efficient}
Tomas Mikolov, Kai Chen, Greg Corrado, and Jeffrey Dean.
\newblock Efficient estimation of word representations in vector space.
\newblock {\em arXiv preprint arXiv:1301.3781}, 2013.

\bibitem[\protect\citeauthoryear{Schlichtkrull \bgroup \em et al.\egroup
  }{2018}]{schlichtkrull2018modeling}
Michael Schlichtkrull, Thomas~N Kipf, Peter Bloem, Rianne Van Den~Berg, Ivan
  Titov, and Max Welling.
\newblock Modeling relational data with graph convolutional networks.
\newblock In {\em European Semantic Web Conference}, pages 593--607. Springer,
  2018.

\bibitem[\protect\citeauthoryear{Sukhbaatar \bgroup \em et al.\egroup
  }{2015}]{sukhbaatar2015end}
Sainbayar Sukhbaatar, Jason Weston, Rob Fergus, et~al.
\newblock End-to-end memory networks.
\newblock In {\em Advances in neural information processing systems}, pages
  2440--2448, 2015.

\bibitem[\protect\citeauthoryear{Xia and Ding}{2019}]{xia-ding-2019-emotion}
Rui Xia and Zixiang Ding.
\newblock Emotion-cause pair extraction: A new task to emotion analysis in
  texts.
\newblock In {\em Proceedings of the 57th Annual Meeting of the Association for
  Computational Linguistics}, pages 1003--1012, Florence, Italy, July 2019.
  Association for Computational Linguistics.

\bibitem[\protect\citeauthoryear{Zhang \bgroup \em et al.\egroup
  }{2019}]{zhang2019syntax}
Chen Zhang, Qiuchi Li, and Dawei Song.
\newblock Syntax-aware aspect-level sentiment classification with
  proximity-weighted convolution network.
\newblock In {\em Proceedings of the 42nd International ACM SIGIR Conference on
  Research and Development in Information Retrieval}, pages 1145--1148, 2019.

\end{thebibliography}

\end{document}